\documentclass[11pt]{article}
\usepackage[a4paper,margin=1in]{geometry}
\usepackage[T1]{fontenc}
\usepackage[utf8]{inputenc}
\usepackage{mathptmx}
\usepackage{amsmath,amssymb,amsthm}
\usepackage{graphicx}
\usepackage{booktabs}
\usepackage{microtype}
\usepackage[round]{natbib}
\usepackage{xcolor}
\usepackage[colorlinks=true,linkcolor=teal!60!black,citecolor=teal!60!black,urlcolor=teal!60!black]{hyperref}
\usepackage{caption}
\usepackage{array}
\newtheorem{proposition}{Proposition}

\newcommand{\br}{[A,B]}
\newcommand{\cnnhead}{\begin{tabular}[b]{@{}c@{}}CIFAR-10 CNN\\[-2pt]{\footnotesize(half learned)}\end{tabular}}
\newcommand{\thA}{\theta_A}

\title{Does Learning Order Explain Forgetting?\\[2pt]
\large The Lie Bracket as an Early-Warning Signal for Catastrophic Forgetting Without Old Data}
\author{Ephraim Kofi Ayi Kwapong\\[2pt]
\small Responsible Artificial Intelligence Lab (RAIL),\\ \small Kwame Nkrumah University of Science and Technology (KNUST), Ghana\\
\small \texttt{ephraimkofi83@gmail.com}}
\date{}

\begin{document}
\maketitle

\begin{abstract}
\noindent
A network trained on task $A$ and then task $B$ ends at different weights than one trained on $B$ and then $A$.
To leading order this gap is set by the Lie bracket of the two tasks' gradient fields, and \citet{dherin2023implicit} conjectured that a non-vanishing bracket may be linked to catastrophic forgetting.
I test this conjecture empirically.
Across Split MNIST, Split Fashion-MNIST and Split CIFAR-10, the forgetting of $A$ is almost exactly the difference in $A$'s loss between the two training orders (Spearman $\rho=0.96$--$1.00$).
I then ask whether the bracket can \emph{predict} that forgetting early in training on $B$ and without any of $A$'s data, using only a small summary stored when $A$ finishes: its gradient and its top 20 Hessian eigenpairs.
When the old task is well learned and forgetting is slow (task-incremental), this stored bracket ranks forgetting with $\rho=0.70$--$0.81$, against $0.33$--$0.58$ for the penalty of Elastic Weight Consolidation, and matches or exceeds re-evaluating $A$ on its own data.
When forgetting is fast (domain-incremental), the stored summary goes stale and the EWC penalty is as good or better.
The bracket fails outright when the old task is only half learned; I trace this to a residual gradient on $A$ an order of magnitude larger, which makes first-order (slope) damage matter.
In five-task sequences with slow forgetting, the stored bracket outperforms the EWC penalty on MNIST and CIFAR-10 features (by $0.16$--$0.17$) and ties it on Fashion-MNIST; it fades as summaries age on MNIST and Fashion-MNIST, where a Taylor estimate from the same summary holds up better.
The bracket is therefore a useful early-warning signal for the quiet kind of forgetting, under a clear condition, provided the summary is kept fresh.
\end{abstract}

\section{Introduction}

Neural networks trained on a sequence of tasks lose performance on earlier ones, a problem known as catastrophic forgetting \citep{mccloskey1989catastrophic,kirkpatrick2017overcoming}.
It matters most for models that keep learning after deployment, such as a crop-disease classifier that is taught a new disease or a clinical model updated with data from a new region.
In those settings the old data is often no longer available, for reasons of privacy, data agreements or storage, so the damage cannot simply be measured by re-testing.

Forgetting comes in two practical forms.
In some settings it is \emph{fast}: accuracy on the old task collapses within a few update steps, and any monitor that looks at the old task catches it.
In others it is \emph{slow}: each update removes a little, the loss is easy to miss, and it accumulates.
The slow case is where an early, data-free warning would be most useful, and it is where existing signals are weakest.

This paper studies one candidate signal.
Gradient descent on two losses does not commute: training on $A$ then $B$ and training on $B$ then $A$ end at different weights.
For small steps the difference is governed by the Lie bracket of the two gradient fields.
\citet{dherin2023implicit} derived this bracket as a correction term in the modified flow of sequential training and conjectured that it may be linked to catastrophic forgetting, but did not test the link.
I test it, and I ask a practical question the conjecture does not address: can the bracket be estimated without the old task's data, and is it then a better predictor of forgetting than the penalty of Elastic Weight Consolidation (EWC), the standard data-free method?

\paragraph{Contributions.}
\begin{itemize}\setlength\itemsep{1pt}
\item \textbf{Order explains forgetting.} In two-task learning on three datasets, the forgetting of $A$ matches the order gap on $A$'s loss almost exactly ($\rho=0.96$--$1.00$; Section~\ref{sec:order}).
\item \textbf{A data-free bracket.} I compute the bracket from a stored summary of $A$ (gradient plus top-$K$ Hessian eigenpairs) and the current task's data. For slow forgetting on well-learned tasks it outperforms the EWC penalty on all three datasets, by $0.23$ to $0.41$ in rank correlation (Section~\ref{sec:slow}).
\item \textbf{An honest boundary.} For fast forgetting the result is mixed, and three attempts to repair the stored bracket did not beat EWC on CIFAR-10 features (Section~\ref{sec:fast}).
\item \textbf{Longer sequences.} In five-task sequences the stored bracket still outperforms the EWC penalty for slow forgetting on MNIST and CIFAR-10 features and ties it on Fashion-MNIST, and its advantage depends on the age of the stored summary (Section~\ref{sec:seq}).
\item \textbf{A measured failure condition.} The bracket fails when the old task is half learned. I show the cause is a residual gradient on $A$ that is an order of magnitude larger relative to $B$'s, and that learning $A$ properly restores the bracket (Section~\ref{sec:mechanism}).
\end{itemize}
I study prediction only. Using the bracket to \emph{reduce} forgetting is left to future work.

\section{Background and related work}

\paragraph{The bracket and order.}
\citet{smith2021origin} and \citet{dherin2023implicit} analyse SGD by backward error analysis.
For two consecutive steps on losses $L_1$ and $L_2$, \citet{dherin2023implicit} shows that the modified flow contains a term proportional to the Lie bracket of the two gradient fields, and conjectures a link to catastrophic forgetting. That work is theoretical and contains no experiments.
The same bracket has been used to study training order: \citet{rukhovich2025commute} use it to analyse how the order of domains affects a target loss in multi-domain learning, and \citet{sweeney2026geometry} uses its projection onto a target gradient to predict which of two orders transfers better.
Neither studies forgetting of an earlier task or compares against EWC.
\citet{bell2022ordering} show empirically that task order changes how much is forgotten.

\paragraph{Second-order views of forgetting.}
EWC \citep{kirkpatrick2017overcoming} penalises movement of weights in proportion to a diagonal Fisher estimate, and Synaptic Intelligence \citep{zenke2017si} accumulates a similar importance online.
\citet{yin2020sola} show that such regularisers can be read as second-order Taylor approximations of the old loss.
\citet{mirzadeh2020understanding} use the same expansion around the old task's minimum to relate forgetting to curvature, dropping the gradient term under the assumption that training has converged.
\citet{yin2020sola} note that the gradient term is usually skipped in practice because it is small at the end of a task, and \citet{wiest2026improved} keeps a first-order term in an EWC-style penalty; both use these approximations as training constraints, not as predictors.
My \emph{stored Taylor estimate} baseline is this expansion with the gradient term kept; Section~\ref{sec:mechanism} shows that the dropped term is exactly what matters when the old task is not well learned.
\citet{doan2021ntk} analyse forgetting through the overlap of task kernels, and \citet{hidekel2026lowrank} give a closed-form predictor of forgetting under a neural-tangent linearisation.
\citet{jin2024forget} forecast which examples a language model forgets after an update; their forecasters rely on cached logits or representations of the old examples, so they need old data.
\citet{ramasesh2021anatomy} study where in the network forgetting arises.

\paragraph{First-order methods.}
GEM and A-GEM \citep{lopezpaz2017gem,chaudhry2019agem}, OGD \citep{farajtabar2020ogd}, GPM \citep{saha2021gpm} and selective projection \citep{singh2026sfao} constrain each update using stored gradients, gradient subspaces or examples.
They mitigate forgetting and use first-order information.
The bracket is a second-order quantity, and here it is used to predict rather than to mitigate.
\citet{chaudhry2019tiny} show that very small replay memories are surprisingly effective, which motivates my tiny-buffer variant.

\section{Method}

\subsection{Setting and the order gap}

Let $L_A$ and $L_B$ be the losses of two tasks, with gradients $g_A,g_B$ and Hessians $H_A,H_B$.
Starting from an initialisation $\theta_0$, I train on $A$ to reach $\thA$, then on $B$ to reach $\theta_{AB}$, using plain SGD with learning rate $\eta$.
The \emph{forgetting} of $A$ is the rise in its test loss,
\begin{equation}
\Delta L_A \;=\; L_A^{\text{test}}(\theta_{AB}) - L_A^{\text{test}}(\thA).
\label{eq:forgetting}
\end{equation}
Training in the opposite order gives $\theta_{BA}$. The \emph{order gap on $A$} is $L_A^{\text{test}}(\theta_{AB}) - L_A^{\text{test}}(\theta_{BA})$.

\begin{proposition}[Order gap of one step]\label{prop:gap}
Let $\theta_{AB}$ be the result of one gradient step on $L_A$ followed by one on $L_B$ from $\theta$, and $\theta_{BA}$ the result of the opposite order. Then
\begin{equation}
\theta_{AB}-\theta_{BA} \;=\; \eta^2\,\br(\theta) + O(\eta^3),
\qquad
\br(\theta) \;=\; H_B(\theta)\,g_A(\theta) - H_A(\theta)\,g_B(\theta).
\label{eq:bracket}
\end{equation}
\end{proposition}
\noindent
The proof is a first-order Taylor expansion of the second gradient in each order.
$\br$ is the Lie bracket of the negative gradient fields $-\nabla L_A$ and $-\nabla L_B$, the quantity in \citet{dherin2023implicit}.
I verified Eq.~\eqref{eq:bracket} numerically on every model: the ratio of measured to predicted gap is between $0.99$ and $1.00$ for $\eta\le 0.01$ (Appendix~\ref{app:checks}).
Proposition~\ref{prop:gap} is exact only for one small step. Whether a quantity defined this way says anything about hundreds of stochastic steps is an empirical question, and it is the subject of this paper.

\subsection{Signals}

All signals are evaluated at the current weights $\theta$ during the first steps of training on $B$.
I use the Euclidean norm of the bracket as a scalar signal. Hessian--vector products are computed by double backpropagation \citep{pearlmutter1994fast}, so no Hessian is ever formed.

\paragraph{Stored summary.}
When training on $A$ ends I store $\mathcal{S}_A=\{\thA,\; g_A(\thA),\; (\lambda_k,u_k)_{k=1}^{K},\; F_A\}$: the weights, the gradient of $A$, the top $K=20$ eigenpairs of $H_A(\thA)$ found by Lanczos iteration, and the diagonal empirical Fisher (used only by the EWC baseline).
Writing $\hat H_A=\sum_k \lambda_k u_k u_k^{\top}$ and $\delta=\theta-\thA$, the summary gives a local model of $A$'s gradient, $\hat g_A(\theta)=g_A(\thA)+\hat H_A\delta$. $A$'s data is then discarded.

\paragraph{Signals that use no old data.}
\begin{align}
\text{Stored bracket (proposed):}\quad & b(\theta)=\big\| H_B(\theta)\,\hat g_A(\theta)-\hat H_A\,g_B(\theta)\big\|, \label{eq:stored}\\
\text{EWC penalty:}\quad & e(\theta)=\textstyle\sum_i F_{A,i}\,\delta_i^2, \label{eq:ewc}\\
\text{Stored Taylor estimate:}\quad & t(\theta)=g_A(\thA)^{\top}\delta+\tfrac12\textstyle\sum_k\lambda_k\,(u_k^{\top}\delta)^2. \label{eq:taylor}
\end{align}
In Eq.~\eqref{eq:stored}, $H_B$ and $g_B$ use the current task's data, which is available by definition.
Eq.~\eqref{eq:ewc} is the quantity EWC adds to the loss \citep{kirkpatrick2017overcoming}; I use it as a predictor, not as a regulariser.
Eq.~\eqref{eq:taylor} estimates the rise in $A$'s loss directly.

\paragraph{Reference signals that use old data.}
The \emph{full bracket} $\|\br(\theta)\|$ uses $A$'s data at the current weights. \emph{Re-evaluation} is the change in $A$'s training loss, $L_A(\theta)-L_A(\thA)$, the most direct monitor when old data is kept.
These are not competitors of the data-free signals. They show what the old data would buy and separate failures of the bracket from failures of the summary.

\subsection{Protocol}

Each signal is recorded at steps $5,10,15,20$ of training on $B$ and averaged. Training on $B$ lasts roughly $370$ to $470$ steps, so the signals see about the first $5\%$.
For every setting I run all $20$ ordered pairs of five tasks with $5$ seeds, giving $100$ runs, and report the Spearman rank correlation $\rho$ between a signal and $\Delta L_A$.
A signal with $\rho=1$ ranks the runs by forgetting perfectly; it need not be calibrated in magnitude.
Confidence intervals are $95\%$ percentile bootstrap intervals over runs ($10{,}000$ resamples), and differences between signals are bootstrapped on the same resamples.
Runs that share a seed or a task are not independent, so the intervals are approximate; I therefore also report per-seed agreement.

\section{Experimental setup}\label{sec:setup}

\begin{table}[t]
\centering\small
\caption{The four setups. Accuracy is the mean test accuracy of a task after it is trained alone. Forgetting is the mean (maximum) drop in $A$'s test accuracy after training on $B$, at $\eta=0.05$.}
\label{tab:setups}
\begin{tabular}{@{}lllccc@{}}
\toprule
Setup & Input & Model & Accuracy & \multicolumn{2}{c}{Forgetting, \% mean (max)}\\
\cmidrule(l){5-6}
 & & & & slow & fast\\
\midrule
MNIST & pixels (784) & MLP 256--256 & 0.989 & 1.3 (7.1) & 35.0 (85.2)\\
Fashion-MNIST & pixels (784) & MLP 256--256 & 0.990 & 0.9 (8.9) & 19.8 (46.5)\\
CIFAR-10 features & frozen ResNet-18 (512) & MLP 256--256 & 0.973 & 0.7 (3.5) & 18.6 (46.2)\\
CIFAR-10 CNN & pixels ($3{\times}32{\times}32$) & small CNN & 0.776 & 4.1 (14.1) & 21.2 (43.4)\\
\bottomrule
\end{tabular}
\end{table}

\paragraph{Tasks.}
I use Split MNIST \citep{lecun1998gradient}, Split Fashion-MNIST \citep{xiao2017fashion} and Split CIFAR-10 \citep{krizhevsky2009learning}. Each dataset is split into five binary tasks by pairing classes in label order (classes $0/1$, $2/3$, and so on).

\paragraph{Two scenarios.}
Following \citet{vandeven2019three}, I use a task-incremental scenario (Task-IL), in which the network has ten outputs and the loss is restricted to the two outputs of the current task, and a domain-incremental scenario (Domain-IL), in which all tasks share the same two outputs.
In my experiments these give slow and fast forgetting respectively (Table~\ref{tab:setups}), and I use the terms interchangeably. The two properties are confounded here; I return to this in Section~\ref{sec:limits}.

\paragraph{Models and training.}
The MLP has two hidden layers of 256 tanh units (about $2.7\times10^5$ parameters on pixels).
For \emph{CIFAR-10 features}, each image is passed once through a ResNet-18 \citep{he2016resnet} pretrained on ImageNet \citep{deng2009imagenet} and frozen; the same MLP is trained on the 512-dimensional output (about $2.0\times10^5$ parameters). Forgetting and all signals are measured on the MLP only.
The \emph{CIFAR-10 CNN} has two $3{\times}3$ convolutional layers (16 and 32 channels, tanh, average pooling) and a 64-unit hidden layer (about $1.4\times10^5$ parameters).
I use tanh and average pooling so that the loss is twice differentiable.
All models are trained with plain SGD without momentum, batch size 64 and learning rate $0.05$ unless stated, for 2 epochs per task (MNIST, Fashion-MNIST) or 3 (CIFAR-10).
I use plain SGD because Proposition~\ref{prop:gap} is derived for memoryless updates.
Gradients, Hessian--vector products and the summary are estimated on a fixed subset of $2{,}000$ training examples per task.

\paragraph{Well-learned and half-learned tasks.}
The first three setups reach $97$--$99\%$ mean accuracy, and at least $94\%$ on every task. The small CNN reaches only $78\%$ and does not improve with twice the training (Appendix~\ref{app:diag}).
I keep it as a deliberate contrast: the same images, with the old task well learned (features) or half learned (CNN).

\section{Results}

\begin{figure}[t]
\centering
\includegraphics[width=\linewidth]{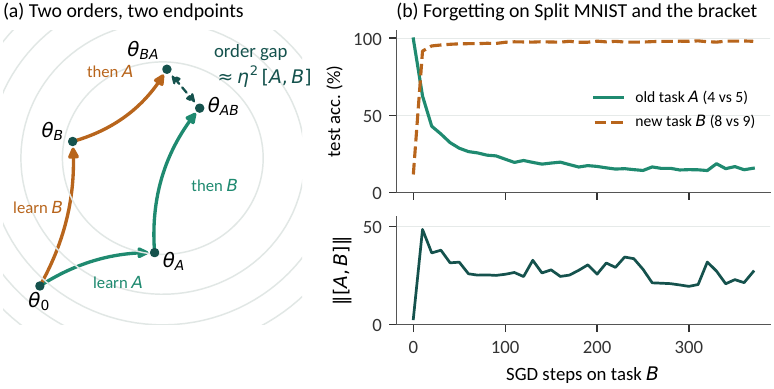}
\caption{(a) Training in the two orders ends at different weights; for one small step the gap is $\eta^2\br$. (b) A single Split MNIST run in the Domain-IL scenario. Accuracy on the old task falls from $99\%$ to about $16\%$ while the new task is learned within tens of steps, and the norm of the bracket jumps at the switch.}
\label{fig:concept}
\end{figure}

\subsection{Forgetting is an order effect}\label{sec:order}

\begin{table}[t]
\centering\small
\caption{Order test ($\eta=0.05$, 3 seeds, 60 runs per cell; slow / fast). Row 1: rank correlation between forgetting $\Delta L_A$ and the order gap on $A$. Row 2: ratio of their mean absolute values. Rows 3--4: rank correlation of the full bracket, averaged over the first 20 steps of $B$, with the order gap on $A$ and with the weight gap $\|\theta_{AB}-\theta_{BA}\|$.}
\label{tab:order}
\begin{tabular}{@{}lcccc@{}}
\toprule
 & MNIST & Fashion-MNIST & CIFAR-10 features & \cnnhead\\
\midrule
Forgetting vs.\ order gap on $A$ & 1.00 / 1.00 & 0.99 / 1.00 & 0.99 / 1.00 & 0.96 / 0.99\\
Order gap $\div$ forgetting & 1.04 / 1.00 & 1.00 / 1.00 & 1.00 / 1.00 & 1.20 / 1.05\\
\addlinespace[2pt]
Bracket vs.\ order gap on $A$ & 0.83 / 0.69 & 0.68 / 0.69 & 0.79 / 0.81 & 0.03 / 0.29\\
Bracket vs.\ weight gap & 0.66 / 0.69 & $-$0.06 / 0.58 & 0.54 / 0.48 & 0.16 / 0.39\\
\bottomrule
\end{tabular}
\end{table}

Table~\ref{tab:order} compares forgetting with the order gap on $A$.
On every setup and in both scenarios the two rank the runs almost identically ($\rho=0.96$--$1.00$) and have nearly the same size (ratio $1.00$--$1.20$).
The explanation is simple: when $A$ is learned last it ends near its own optimum, so the difference between the two orders is the damage done to $A$ by learning $B$ after it.
A ratio above one means that learning $A$ after $B$ ends slightly better than learning $A$ alone.
This result is close to true by construction, and I report it for that reason: it shows that in this setting predicting forgetting and predicting the order effect are the same problem.
The remaining question is whether the bracket, which describes the order effect for one step, tracks it over a whole training run.
Rows 3--4 show that it does when the old task is well learned ($\rho=0.68$--$0.83$ with the order gap on $A$) and does not on the half-learned CNN ($0.03$ and $0.29$).

\subsection{Slow forgetting: the stored bracket outperforms EWC without old data}\label{sec:slow}

\begin{figure}[t]
\centering
\includegraphics[width=\linewidth]{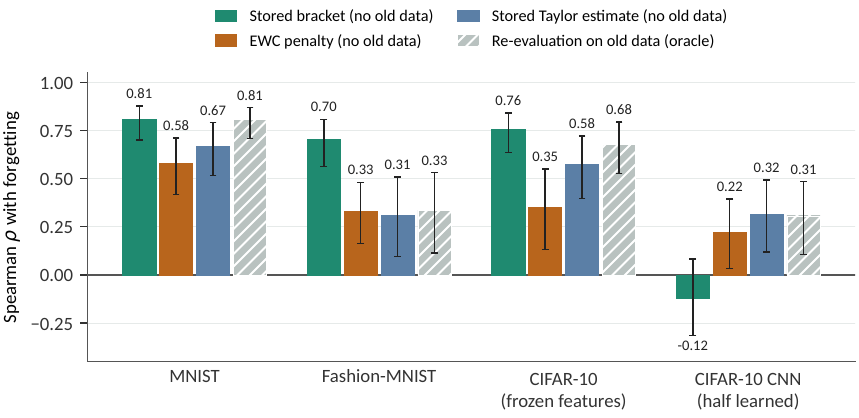}
\caption{Slow forgetting (Task-IL), $\eta=0.05$, 100 runs per setup. Rank correlation of each signal with forgetting; bars show 95\% bootstrap intervals. The first three signals use no old data. The hatched bar re-evaluates the old task on its own data and is shown for reference.}
\label{fig:slow}
\end{figure}

\begin{table}[t]
\centering\small
\caption{Rank correlation with forgetting, $\eta=0.05$, 100 runs per cell, with 95\% bootstrap intervals. The best data-free signal in each column is in bold. $^\dagger$Reference signals that use the old task's data.}
\label{tab:main}
\setlength{\tabcolsep}{4pt}
\begin{tabular}{@{}lcccc@{}}
\toprule
 & MNIST & Fashion-MNIST & CIFAR-10 features & \cnnhead\\
\midrule
\multicolumn{5}{l}{\emph{Slow forgetting (Task-IL)}}\\
Stored bracket (proposed) & \textbf{0.81} {\scriptsize[0.70, 0.88]} & \textbf{0.70} {\scriptsize[0.56, 0.81]} & \textbf{0.76} {\scriptsize[0.64, 0.84]} & $-$0.12 {\scriptsize[$-$0.31, 0.08]}\\
EWC penalty & 0.58 {\scriptsize[0.42, 0.71]} & 0.33 {\scriptsize[0.16, 0.48]} & 0.35 {\scriptsize[0.13, 0.55]} & 0.22 {\scriptsize[0.03, 0.39]}\\
Stored Taylor estimate & 0.67 {\scriptsize[0.52, 0.79]} & 0.31 {\scriptsize[0.10, 0.51]} & 0.58 {\scriptsize[0.40, 0.72]} & \textbf{0.32} {\scriptsize[0.12, 0.49]}\\
\addlinespace[2pt]
Full bracket$^\dagger$ & 0.82 {\scriptsize[0.72, 0.88]} & 0.74 {\scriptsize[0.61, 0.84]} & 0.76 {\scriptsize[0.65, 0.84]} & $-$0.11 {\scriptsize[$-$0.30, 0.09]}\\
Re-evaluation$^\dagger$ & 0.81 {\scriptsize[0.71, 0.87]} & 0.33 {\scriptsize[0.11, 0.53]} & 0.68 {\scriptsize[0.53, 0.79]} & 0.31 {\scriptsize[0.11, 0.49]}\\
\addlinespace[2pt]
Bracket $-$ EWC & $+0.23$ {\scriptsize[0.13, 0.33]} & $+0.37$ {\scriptsize[0.21, 0.53]} & $+0.41$ {\scriptsize[0.21, 0.60]} & $-0.34$ {\scriptsize[$-$0.53, $-$0.15]}\\
\midrule
\multicolumn{5}{l}{\emph{Fast forgetting (Domain-IL)}}\\
Stored bracket (proposed) & 0.40 {\scriptsize[0.22, 0.56]} & 0.78 {\scriptsize[0.68, 0.84]} & 0.46 {\scriptsize[0.28, 0.61]} & 0.30 {\scriptsize[0.09, 0.49]}\\
EWC penalty & \textbf{0.59} {\scriptsize[0.41, 0.74]} & \textbf{0.79} {\scriptsize[0.70, 0.85]} & \textbf{0.84} {\scriptsize[0.76, 0.88]} & 0.47 {\scriptsize[0.30, 0.62]}\\
Stored Taylor estimate & 0.41 {\scriptsize[0.20, 0.58]} & 0.78 {\scriptsize[0.68, 0.85]} & 0.71 {\scriptsize[0.59, 0.79]} & \textbf{0.76} {\scriptsize[0.66, 0.84]}\\
\addlinespace[2pt]
Full bracket$^\dagger$ & 0.69 {\scriptsize[0.56, 0.79]} & 0.67 {\scriptsize[0.55, 0.77]} & 0.81 {\scriptsize[0.73, 0.87]} & 0.29 {\scriptsize[0.09, 0.47]}\\
Re-evaluation$^\dagger$ & 0.81 {\scriptsize[0.73, 0.87]} & 0.81 {\scriptsize[0.71, 0.87]} & 0.84 {\scriptsize[0.76, 0.89]} & 0.78 {\scriptsize[0.68, 0.85]}\\
\addlinespace[2pt]
Bracket $-$ EWC & $-0.19$ {\scriptsize[$-$0.33, $-$0.06]} & $-0.01$ {\scriptsize[$-$0.10, 0.07]} & $-0.38$ {\scriptsize[$-$0.55, $-$0.22]} & $-0.17$ {\scriptsize[$-$0.37, 0.03]}\\
\bottomrule
\end{tabular}
\end{table}

Figure~\ref{fig:slow} and Table~\ref{tab:main} (top) give the main result.
On the three well-learned setups the stored bracket ranks forgetting with $\rho=0.81$, $0.70$ and $0.76$, against $0.58$, $0.33$ and $0.35$ for the EWC penalty.
The differences, $+0.23$, $+0.37$ and $+0.41$, have bootstrap intervals that exclude zero, and the bracket has the higher correlation in all 15 seed-level comparisons (5 seeds on each of 3 datasets).
The stored bracket is also as good as the full bracket computed with old data ($0.82$, $0.74$, $0.76$), so in this regime the 20-direction summary loses little.

Two comparisons deserve care.
First, against the stored Taylor estimate, which uses the same summary, the bracket is higher on all three setups ($0.81$ vs.\ $0.67$, $0.70$ vs.\ $0.31$, $0.76$ vs.\ $0.58$); the difference is clear on Fashion-MNIST ($+0.39$, interval $[0.23,0.56]$) and CIFAR-10 features ($+0.18$, $[0.03,0.34]$) and within noise on MNIST ($+0.14$, $[-0.02,0.30]$).
Second, against re-evaluation on the old data, the bracket is tied on MNIST ($0.81$ vs.\ $0.81$), higher within noise on CIFAR-10 features ($0.76$ vs.\ $0.68$; difference $+0.08$, interval $[-0.05,0.21]$) and clearly higher on Fashion-MNIST ($0.70$ vs.\ $0.33$; $+0.37$, $[0.21,0.55]$).
So a signal that uses no old data matches or exceeds direct monitoring with all of it.
A likely reason is that slow forgetting has barely begun after 20 steps, so the change in $A$'s loss is still small relative to noise, whereas the bracket measures a rate.

Forgetting in this scenario is small (Table~\ref{tab:setups}), which is the defining property of the slow case. The correlations are computed over 100 runs and are stable across seeds, but the absolute stakes per update are low.
The result holds across learning rates on MNIST and is strongest for small steps ($0.86$, $0.81$, $0.71$ at $\eta=0.02,0.05,0.1$, against $0.55$--$0.59$ for EWC; Appendix~\ref{app:lr}), as expected from a small-step expansion.

\subsection{Fast forgetting: the summary goes stale}\label{sec:fast}

\begin{figure}[t]
\centering
\includegraphics[width=\linewidth]{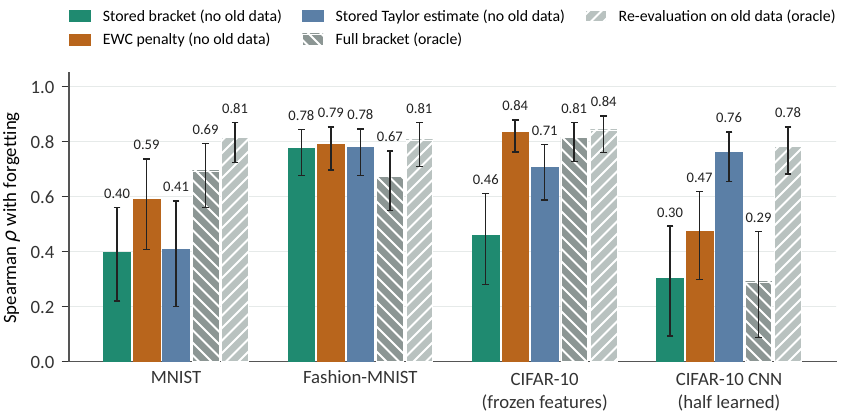}
\caption{Fast forgetting (Domain-IL), $\eta=0.05$, 100 runs per setup. Hatched bars are reference signals that use old data.}
\label{fig:fast}
\end{figure}

For fast forgetting the picture changes (Figure~\ref{fig:fast}, Table~\ref{tab:main} bottom).
Without old data, the EWC penalty is better than the stored bracket on MNIST ($0.59$ vs.\ $0.40$) and CIFAR-10 features ($0.84$ vs.\ $0.46$) and tied with it on Fashion-MNIST ($0.79$ vs.\ $0.78$).

The full bracket remains a good predictor on MNIST and CIFAR-10 features ($0.69$ and $0.81$), so the bracket itself has not failed. The stored version has.
In fast forgetting the weights move far from $\thA$ within a few steps, and a local model of $A$ built at $\thA$ no longer describes $A$ where the network now is.
The EWC penalty is a function of distance alone, and it continues to track the damage.
Re-evaluation on old data is the strongest signal, or tied for it, in every fast setting ($0.78$--$0.84$): fast forgetting is easy to detect when it can be observed.

\paragraph{A tiny buffer helps on MNIST only.}
If ten examples of $A$ are kept and the bracket is computed from them instead of from the summary, it overtakes EWC on MNIST at all three learning rates ($0.74$, $0.72$, $0.57$ vs.\ $0.61$, $0.59$, $0.49$; Appendix~\ref{app:buffer}).
This does not carry over to CIFAR-10 features, where ten examples give $0.28$ and fifty give $0.69$, still below EWC.

\paragraph{Repairs that did not work.}
On CIFAR-10 features I tried three ways to repair the data-free bracket in fast forgetting (Appendix~\ref{app:fixes}): adding its rank to that of EWC ($0.71$), adding its rank to that of the Taylor estimate ($0.62$), and enlarging the summary to $K=50$ or $100$ eigenpairs ($0.49$ in both cases, 3 seeds).
None reaches the EWC penalty alone ($0.84$).
The size of the summary is therefore not the limiting factor; its locality is.

\subsection{When the bracket fails: half-learned tasks}\label{sec:mechanism}

\begin{figure}[t]
\centering
\includegraphics[width=\linewidth]{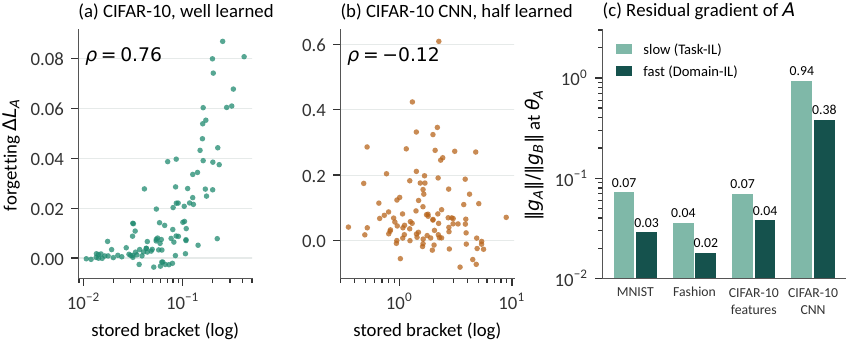}
\caption{(a,b) Stored bracket against forgetting for the same CIFAR-10 tasks in the slow scenario, with the old task well learned (frozen features, $97\%$ accuracy) and half learned (small CNN, $78\%$). (c) Norm of the old task's gradient at $\thA$ relative to the new task's.}
\label{fig:mechanism}
\end{figure}

\begin{table}[t]
\centering\small
\caption{Mechanism test ($\eta=0.05$, 3 seeds; slow / fast). Slope share is $|g_A^{\top}\delta| / (|g_A^{\top}\delta| + |\tfrac12\delta^{\top}H_A\delta|)$ at step 20, with means taken over runs. The last four rows are rank correlations with forgetting.}
\label{tab:mech}
\setlength{\tabcolsep}{5pt}
\begin{tabular}{@{}lcccc@{}}
\toprule
 & MNIST & Fashion-MNIST & CIFAR-10 features & \cnnhead\\
\midrule
$\|g_A\|/\|g_B\|$ at $\thA$ & 0.07 / 0.03 & 0.04 / 0.02 & 0.07 / 0.04 & 0.94 / 0.38\\
Slope share of damage & 0.31 / 0.11 & 0.35 / 0.12 & 0.25 / 0.10 & 0.57 / 0.23\\
\addlinespace[2pt]
Full bracket & 0.82 / 0.69 & 0.67 / 0.69 & 0.78 / 0.80 & 0.04 / 0.30\\
\quad term $\|H_A g_B\|$ alone & 0.78 / 0.79 & 0.71 / 0.77 & 0.79 / 0.75 & 0.02 / 0.19\\
\quad term $\|H_B g_A\|$ alone & 0.75 / 0.68 & 0.50 / 0.63 & 0.68 / 0.72 & 0.04 / 0.25\\
Measured curvature damage & 0.84 / 0.47 & 0.71 / 0.80 & 0.87 / 0.78 & 0.17 / 0.73\\
Measured slope damage & 0.01 / 0.37 & $-$0.29 / $-$0.55 & 0.04 / 0.10 & 0.43 / 0.00\\
\bottomrule
\end{tabular}
\end{table}

On the CIFAR-10 CNN the bracket does not predict forgetting in either scenario ($-0.12$ and $0.30$ stored; $-0.11$ and $0.29$ with old data), so the summary is not at fault.
Figure~\ref{fig:mechanism}(a,b) shows the contrast on the same images.

The difference is how well $A$ has been learned.
Expanding $A$'s loss around $\thA$,
\begin{equation}
L_A(\theta)-L_A(\thA) \;\approx\; \underbrace{g_A(\thA)^{\top}\delta}_{\text{slope}} \;+\; \underbrace{\tfrac12\,\delta^{\top}H_A(\thA)\,\delta}_{\text{curvature}},
\label{eq:split}
\end{equation}
the first term vanishes only if $A$ has converged.
Table~\ref{tab:mech} measures each part. On the well-learned setups the residual gradient of $A$ is $2$--$7\%$ of $B$'s. On the CNN it is $94\%$ (slow) and $38\%$ (fast), an order of magnitude larger.
In slow forgetting on the CNN the slope term accounts for $57\%$ of the damage and is the better predictor of forgetting ($0.43$ against $0.17$ for curvature).
The bracket contains $H_A g_B$, a curvature quantity, and has no term that responds to movement along $A$'s residual gradient, so it misses this.

Learning $A$ properly removes the problem.
With frozen pretrained features, every measure in Table~\ref{tab:mech} returns to the MNIST pattern: the residual gradient is small, damage is mostly curvature, the term $\|H_A g_B\|$ alone predicts forgetting ($0.79$ and $0.75$), and the stored bracket goes from $-0.12$ to $0.76$ in the slow scenario.
Longer training could not be used to test this on the CNN, whose accuracy does not improve with more epochs.

\paragraph{What remains unexplained.}
In fast forgetting on the CNN the damage is still mostly curvature ($77\%$), and the measured curvature term predicts forgetting well ($0.73$), yet $\|H_A g_B\|$ does not ($0.19$).
I tested whether $B$'s gradient points into directions of negative curvature of $A$, which the norm would misread. It does not: the cosine between $g_B$ and $H_A g_B$ is $0.82$ on the CNN, higher than on the well-learned setups ($0.59$--$0.62$), although the CNN does have more negative curvature (Appendix~\ref{app:valley}).
The CNN's largest curvature is three to seven times larger than in the other setups, and one possibility is that its Hessian changes quickly along the path, so that curvature measured at the current point no longer describes the accumulated damage. I leave this open.

In all three well-learned setups the single term $\|H_A g_B\|$ is about as good as the full bracket, and better in fast forgetting on MNIST and Fashion-MNIST. The predictive content of the bracket lies mainly in how $A$'s curvature acts on $B$'s gradient.

\subsection{Longer task sequences}\label{sec:seq}

\begin{figure}[t]
\centering
\includegraphics[width=\linewidth]{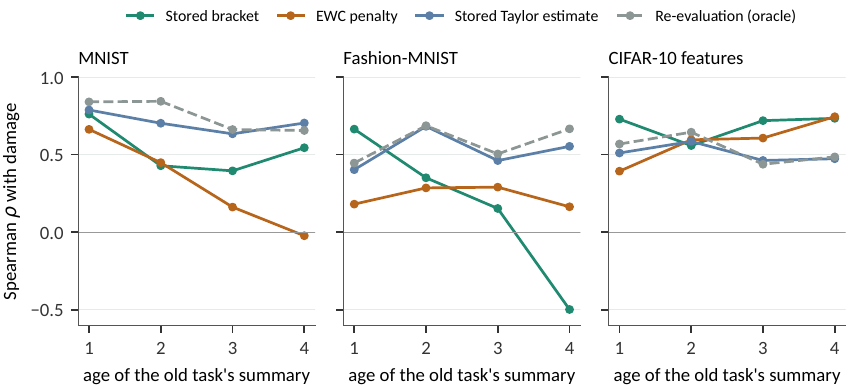}
\caption{Five-task sequences, slow forgetting (Task-IL): rank correlation with the damage a new task does to an earlier task, by the age of the earlier task's stored summary. The fast scenario is in Figure~\ref{fig:seq_fast}.}
\label{fig:seq}
\end{figure}

\begin{table}[t]
\centering\small
\caption{Five-task sequences: rank correlation between each signal and the damage a new task does to each earlier task (200 events per cell: 5 seeds $\times$ 4 task orders $\times$ 10 old/new pairs, $\eta=0.05$). The best data-free signal in each column is in bold. Intervals are $95\%$ bootstrap intervals over whole sequences. $^\dagger$Reference signals that use the old task's data.}
\label{tab:seq}
\resizebox{\linewidth}{!}{\begin{tabular}{@{}lcccccc@{}}
\toprule
 & \multicolumn{3}{c}{Slow forgetting (Task-IL)} & \multicolumn{3}{c}{Fast forgetting (Domain-IL)}\\
\cmidrule(lr){2-4}\cmidrule(l){5-7}
 & MNIST & Fashion & CIFAR feat. & MNIST & Fashion & CIFAR feat.\\
\midrule
Stored bracket (proposed) & 0.58 & 0.27 & \textbf{0.68} & 0.16 & 0.30 & $-$0.03\\
EWC penalty & 0.41 & 0.23 & 0.53 & \textbf{0.69} & 0.43 & \textbf{0.76}\\
Stored Taylor estimate & \textbf{0.71} & \textbf{0.57} & 0.51 & 0.50 & \textbf{0.60} & 0.62\\
Full bracket$^\dagger$ & 0.59 & 0.15 & 0.65 & 0.13 & 0.12 & 0.06\\
Re-evaluation$^\dagger$ & 0.78 & 0.62 & 0.53 & 0.90 & 0.62 & 0.86\\
\midrule
Bracket $-$ EWC, all ages & +0.17 {\scriptsize[0.06, 0.28]} & +0.04 {\scriptsize[$-$0.07, 0.16]} & +0.16 {\scriptsize[0.07, 0.25]} & $-$0.53 {\scriptsize[$-$0.61, $-$0.44]} & $-$0.13 {\scriptsize[$-$0.29, 0.02]} & $-$0.78 {\scriptsize[$-$0.93, $-$0.61]}\\
Bracket $-$ EWC, age 1 & +0.10 {\scriptsize[$-$0.04, 0.20]} & +0.48 {\scriptsize[0.32, 0.64]} & +0.34 {\scriptsize[0.09, 0.60]} & $-$0.25 {\scriptsize[$-$0.42, $-$0.10]} & +0.12 {\scriptsize[0.04, 0.21]} & $-$0.45 {\scriptsize[$-$0.65, $-$0.26]}\\
\bottomrule
\end{tabular}}
\end{table}

All results so far are for a single pair $A\to B$. Here each run learns all five tasks of a dataset one after another, in a random order, with $4$ orders for each of $5$ seeds ($20$ sequences per setting).
When a task finishes, its summary $\mathcal{S}$ is stored and its data is not used again.
During the first steps of every later task $k$, each signal is computed for every earlier task $j$ from $j$'s summary, and compared with the rise in $j$'s test loss during task $k$, the damage done by $k$.
This gives $10$ old/new events per sequence and $200$ per setting.
The \emph{age} of an event is the number of tasks learned since $j$'s summary was stored ($1$ to $4$).
Because $j$ may already have drifted before $k$ starts, the EWC penalty and the Taylor estimate are measured as their growth since $k$ started; using their totals since $j$ ended instead gives lower correlations in every slow setting (Appendix~\ref{app:seq}).
Intervals are bootstrapped over whole sequences, which keeps the events of one sequence together.

\paragraph{Slow forgetting.}
Over all ages, the stored bracket outperforms the EWC penalty on MNIST ($+0.17$) and CIFAR-10 features ($+0.16$), with intervals that exclude zero, and ties it on Fashion-MNIST ($+0.04$, interval includes zero; Table~\ref{tab:seq}).
With a fresh summary (age $1$) the bracket's lead is largest: $+0.48$ on Fashion-MNIST and $+0.34$ on CIFAR-10 features.
As the summary ages, the bracket fades on MNIST and Fashion-MNIST but not on CIFAR-10 features, where it stays at $0.72$--$0.73$ at ages $3$ and $4$ (Figure~\ref{fig:seq}).
The full bracket, computed with the old task's data at the current weights, fades in the same way, so the loss is not only due to a stale summary.
On MNIST and Fashion-MNIST the stored Taylor estimate, built from the same summary, is the best data-free signal overall ($0.71$ and $0.57$) and holds up better with age. On CIFAR-10 features it does not ($0.51$).

\paragraph{Fast forgetting.}
The pairwise picture carries over. The EWC penalty is the best data-free signal on MNIST and CIFAR-10 features and the Taylor estimate on Fashion-MNIST; the bracket is informative only at age $1$ and is near zero or negative for older summaries.

\paragraph{Drift.}
Old tasks drift before the new task begins: in the slow sequences their test loss has already risen by $0.03$--$0.10$ (MNIST), $0.03$--$0.09$ (Fashion-MNIST) and $0.02$--$0.05$ (CIFAR-10 features) at ages $2$ to $4$; from age $3$ on, this drift is larger than the damage of a single new task (Appendix~\ref{app:seq}, Table~\ref{tab:drift}).
CIFAR-10 features, where the bracket does not fade, drifts about half as much as the other two. This is consistent with the bracket losing its grip as an old task moves away from the point where its summary was taken, as in the half-learned case of Section~\ref{sec:mechanism}, but the evidence is indirect, and Fashion-MNIST fades more than MNIST despite drifting slightly less.

The practical reading is that the summary should be refreshed: the bracket is most reliable when the summary describes the old task near its current weights.

\section{Discussion and limitations}\label{sec:limits}

\paragraph{What the bracket is for.}
The results suggest a division of labour.
Fast forgetting is loud: the EWC penalty detects it without data and re-evaluation detects it with data.
Slow forgetting is quiet, and there, in two-task learning, the stored bracket is the best data-free signal I tested and is competitive with re-evaluation on all old data. Over five-task sequences it keeps its lead over EWC on two of three datasets, but a Taylor estimate from the same summary does better on two of the three datasets.
The bracket is therefore best seen as an early warning for the kind of forgetting that is otherwise easy to miss, in settings where the old data cannot be kept.

\paragraph{Limitations.}
\begin{itemize}\setlength\itemsep{1pt}
\item \emph{Scale.} The models are small MLPs and one small CNN; on CIFAR-10 the positive result is for a frozen pretrained backbone with a trainable head.
\item \emph{Short sequences.} The longest sequences have five tasks, so summaries are at most four tasks old, and the age-4 groups contain only $20$ events.
\item \emph{Scenario and speed are confounded.} Slow forgetting is Task-IL and fast forgetting is Domain-IL in all experiments.
\item \emph{Small effects in the slow case.} Mean forgetting there is about $1\%$ of accuracy.
\item \emph{Easy pairs.} Two of the five Fashion-MNIST pairs reach $100\%$ accuracy, so that dataset confirms the pattern more than it extends it.
\item \emph{Ranking only.} I report rank correlations; the signals are not calibrated estimates of the loss.
\item \emph{Optimiser.} All results use plain SGD. Momentum and Adam would need a modified bracket.
\item \emph{Statistics.} Main results use 5 seeds; the order and mechanism tests use 3, the curvature test 2. Bootstrap intervals for pairs treat runs as independent; for sequences they resample whole sequences.
\item \emph{Cost.} The summary stores $(K{+}2)P$ numbers for a model with $P$ parameters, against $2P$ for EWC, and each evaluation of the bracket needs one Hessian--vector product on the current task.
\end{itemize}

\paragraph{Future work.}
The natural next step is to use the bracket to reduce forgetting, by steering updates away from directions in which $H_A g_B$ is large, and to compare against EWC, replay and first-order projection methods \citep{farajtabar2020ogd,saha2021gpm} over longer task sequences.
The sequence results suggest refreshing the summary as tasks accumulate, and combining the bracket with the Taylor estimate from the same summary.
Explaining the failure of the curvature term in fast forgetting on the CNN, and finding a summary that stays valid far from $\thA$, are open problems.

\section{Conclusion}

I tested the conjecture that the Lie bracket of task gradients is linked to catastrophic forgetting.
In two-task learning, forgetting is an order effect, and the bracket tracks it when the old task is well learned.
Computed from a small stored summary and no old data, the bracket predicts slow forgetting better than the EWC penalty on three datasets.
It does not beat EWC for fast forgetting, and it fails when the old task has a large residual gradient, for a reason I measured.
In five-task sequences it still outperforms the EWC penalty for slow forgetting on two of three datasets and ties it on the third, and its advantage depends on how fresh the stored summary is.
These boundaries make the claim narrower than the conjecture, and more useful: the bracket is a practical early warning for quiet forgetting on well-trained models, provided its summary is kept fresh.

\section*{Use of AI tools}
An AI assistant (Claude, Anthropic) was used to help write the analysis code, produce figures and draft parts of the text. The author designed the study, ran all experiments, verified the results and references, and takes full responsibility for the content.

\section*{Code and data}
The five notebooks used for all experiments (including \texttt{forgetting\_sequence.ipynb} for Section~\ref{sec:seq}), together with the per-run result files behind every table and figure, are available at \url{https://github.com/Hackme262/lie-bracket-forgetting}.

\bibliographystyle{plainnat}
\bibliography{refs}

\begin{thebibliography}{28}
\providecommand{\natexlab}[1]{#1}
\providecommand{\url}[1]{\texttt{#1}}
\expandafter\ifx\csname urlstyle\endcsname\relax
  \providecommand{\doi}[1]{doi: #1}\else
  \providecommand{\doi}{doi: \begingroup \urlstyle{rm}\Url}\fi

\bibitem[Bell and Lawrence(2022)]{bell2022ordering}
Samuel~J. Bell and Neil~D. Lawrence.
\newblock The effect of task ordering in continual learning.
\newblock \emph{arXiv preprint arXiv:2205.13323}, 2022.

\bibitem[Chaudhry et~al.(2019{\natexlab{a}})Chaudhry, Ranzato, Rohrbach, and
  Elhoseiny]{chaudhry2019agem}
Arslan Chaudhry, Marc'Aurelio Ranzato, Marcus Rohrbach, and Mohamed Elhoseiny.
\newblock Efficient lifelong learning with {A-GEM}.
\newblock In \emph{International Conference on Learning Representations
  (ICLR)}, 2019{\natexlab{a}}.

\bibitem[Chaudhry et~al.(2019{\natexlab{b}})Chaudhry, Rohrbach, Elhoseiny,
  Ajanthan, Dokania, Torr, and Ranzato]{chaudhry2019tiny}
Arslan Chaudhry, Marcus Rohrbach, Mohamed Elhoseiny, Thalaiyasingam Ajanthan,
  Puneet~K. Dokania, Philip H.~S. Torr, and Marc'Aurelio Ranzato.
\newblock On tiny episodic memories in continual learning.
\newblock \emph{arXiv preprint arXiv:1902.10486}, 2019{\natexlab{b}}.

\bibitem[Deng et~al.(2009)Deng, Dong, Socher, Li, Li, and
  Fei-Fei]{deng2009imagenet}
Jia Deng, Wei Dong, Richard Socher, Li-Jia Li, Kai Li, and Li~Fei-Fei.
\newblock {ImageNet}: A large-scale hierarchical image database.
\newblock In \emph{Proceedings of the IEEE Conference on Computer Vision and
  Pattern Recognition (CVPR)}, pages 248--255, 2009.

\bibitem[Dherin(2023)]{dherin2023implicit}
Benoit Dherin.
\newblock Implicit biases in multitask and continual learning from a backward
  error analysis perspective.
\newblock In \emph{NeurIPS 2023 Workshop on Mathematics of Modern Machine
  Learning (M3L)}, 2023.
\newblock arXiv:2311.00235.

\bibitem[Doan et~al.(2021)Doan, Bennani, Mazoure, Rabusseau, and
  Alquier]{doan2021ntk}
Thang Doan, Mehdi~Abbana Bennani, Bogdan Mazoure, Guillaume Rabusseau, and
  Pierre Alquier.
\newblock A theoretical analysis of catastrophic forgetting through the {NTK}
  overlap matrix.
\newblock In \emph{Proceedings of the 24th International Conference on
  Artificial Intelligence and Statistics (AISTATS)}, volume 130 of \emph{PMLR},
  pages 1072--1080, 2021.

\bibitem[Farajtabar et~al.(2020)Farajtabar, Azizan, Mott, and
  Li]{farajtabar2020ogd}
Mehrdad Farajtabar, Navid Azizan, Alex Mott, and Ang Li.
\newblock Orthogonal gradient descent for continual learning.
\newblock In \emph{Proceedings of the 23rd International Conference on
  Artificial Intelligence and Statistics (AISTATS)}, volume 108 of \emph{PMLR},
  pages 3762--3773, 2020.
\newblock arXiv:1910.07104.

\bibitem[He et~al.(2016)He, Zhang, Ren, and Sun]{he2016resnet}
Kaiming He, Xiangyu Zhang, Shaoqing Ren, and Jian Sun.
\newblock Deep residual learning for image recognition.
\newblock In \emph{Proceedings of the IEEE Conference on Computer Vision and
  Pattern Recognition (CVPR)}, pages 770--778, 2016.

\bibitem[Hidekel and Raviv(2026)]{hidekel2026lowrank}
Ido~Nitzan Hidekel and Dan Raviv.
\newblock Catastrophic forgetting is low-rank: A function-space theory for
  continual adaptation.
\newblock \emph{arXiv preprint arXiv:2606.18024}, 2026.

\bibitem[Jin and Ren(2024)]{jin2024forget}
Xisen Jin and Xiang Ren.
\newblock What will my model forget? forecasting forgotten examples in language
  model refinement.
\newblock In \emph{Proceedings of the 41st International Conference on Machine
  Learning (ICML)}, volume 235 of \emph{PMLR}, pages 22145--22159, 2024.

\bibitem[Kirkpatrick et~al.(2017)Kirkpatrick, Pascanu, Rabinowitz, Veness,
  Desjardins, Rusu, Milan, Quan, Ramalho, Grabska-Barwinska, Hassabis, Clopath,
  Kumaran, and Hadsell]{kirkpatrick2017overcoming}
James Kirkpatrick, Razvan Pascanu, Neil Rabinowitz, Joel Veness, Guillaume
  Desjardins, Andrei~A. Rusu, Kieran Milan, John Quan, Tiago Ramalho, Agnieszka
  Grabska-Barwinska, Demis Hassabis, Claudia Clopath, Dharshan Kumaran, and
  Raia Hadsell.
\newblock Overcoming catastrophic forgetting in neural networks.
\newblock \emph{Proceedings of the National Academy of Sciences}, 114\penalty0
  (13):\penalty0 3521--3526, 2017.

\bibitem[Krizhevsky(2009)]{krizhevsky2009learning}
Alex Krizhevsky.
\newblock Learning multiple layers of features from tiny images.
\newblock Technical report, University of Toronto, 2009.

\bibitem[LeCun et~al.(1998)LeCun, Bottou, Bengio, and
  Haffner]{lecun1998gradient}
Yann LeCun, L{\'e}on Bottou, Yoshua Bengio, and Patrick Haffner.
\newblock Gradient-based learning applied to document recognition.
\newblock \emph{Proceedings of the IEEE}, 86\penalty0 (11):\penalty0
  2278--2324, 1998.

\bibitem[Lopez-Paz and Ranzato(2017)]{lopezpaz2017gem}
David Lopez-Paz and Marc'Aurelio Ranzato.
\newblock Gradient episodic memory for continual learning.
\newblock In \emph{Advances in Neural Information Processing Systems
  (NeurIPS)}, pages 6467--6476, 2017.

\bibitem[McCloskey and Cohen(1989)]{mccloskey1989catastrophic}
Michael McCloskey and Neal~J. Cohen.
\newblock Catastrophic interference in connectionist networks: The sequential
  learning problem.
\newblock In \emph{Psychology of Learning and Motivation}, volume~24, pages
  109--165. Academic Press, 1989.

\bibitem[Mirzadeh et~al.(2020)Mirzadeh, Farajtabar, Pascanu, and
  Ghasemzadeh]{mirzadeh2020understanding}
Seyed~Iman Mirzadeh, Mehrdad Farajtabar, Razvan Pascanu, and Hassan
  Ghasemzadeh.
\newblock Understanding the role of training regimes in continual learning.
\newblock In \emph{Advances in Neural Information Processing Systems
  (NeurIPS)}, 2020.

\bibitem[Pearlmutter(1994)]{pearlmutter1994fast}
Barak~A. Pearlmutter.
\newblock Fast exact multiplication by the {H}essian.
\newblock \emph{Neural Computation}, 6\penalty0 (1):\penalty0 147--160, 1994.

\bibitem[Ramasesh et~al.(2021)Ramasesh, Dyer, and Raghu]{ramasesh2021anatomy}
Vinay~V. Ramasesh, Ethan Dyer, and Maithra Raghu.
\newblock Anatomy of catastrophic forgetting: Hidden representations and task
  semantics.
\newblock In \emph{International Conference on Learning Representations
  (ICLR)}, 2021.
\newblock arXiv:2007.07400.

\bibitem[Rukhovich et~al.(2025)Rukhovich, Podolskiy, and
  Piontkovskaya]{rukhovich2025commute}
Alexey Rukhovich, Alexander Podolskiy, and Irina Piontkovskaya.
\newblock Commute your domains: Trajectory optimality criterion for
  multi-domain learning.
\newblock \emph{arXiv preprint arXiv:2501.15556}, 2025.

\bibitem[Saha et~al.(2021)Saha, Garg, and Roy]{saha2021gpm}
Gobinda Saha, Isha Garg, and Kaushik Roy.
\newblock Gradient projection memory for continual learning.
\newblock In \emph{International Conference on Learning Representations
  (ICLR)}, 2021.

\bibitem[Singh et~al.(2026)Singh, Dhaulakhandi, Chopade, Malipati, Martinez,
  and Zhu]{singh2026sfao}
Anika Singh, Aayush Dhaulakhandi, Varun Chopade, Likhith Malipati, David
  Martinez, and Kevin Zhu.
\newblock Mitigating forgetting in continual learning with selective gradient
  projection.
\newblock \emph{arXiv preprint arXiv:2603.26671}, 2026.

\bibitem[Smith et~al.(2021)Smith, Dherin, Barrett, and De]{smith2021origin}
Samuel~L. Smith, Benoit Dherin, David G.~T. Barrett, and Soham De.
\newblock On the origin of implicit regularization in stochastic gradient
  descent.
\newblock In \emph{International Conference on Learning Representations
  (ICLR)}, 2021.

\bibitem[Sweeney(2026)]{sweeney2026geometry}
John Sweeney.
\newblock The geometry of sequential learning: {L}ie-bracket prediction of
  transfer order.
\newblock In \emph{Proceedings of the 43rd International Conference on Machine
  Learning (ICML)}, volume 306 of \emph{PMLR}, pages 117806--117825, 2026.
\newblock arXiv:2606.24993.

\bibitem[van~de Ven and Tolias(2019)]{vandeven2019three}
Gido~M. van~de Ven and Andreas~S. Tolias.
\newblock Three scenarios for continual learning.
\newblock \emph{arXiv preprint arXiv:1904.07734}, 2019.

\bibitem[Wiest(2026)]{wiest2026improved}
Davide Wiest.
\newblock Improved elastic weight consolidation as an optimization constraint
  for continual learning.
\newblock \emph{arXiv preprint arXiv:2611.16039}, 2026.

\bibitem[Xiao et~al.(2017)Xiao, Rasul, and Vollgraf]{xiao2017fashion}
Han Xiao, Kashif Rasul, and Roland Vollgraf.
\newblock Fashion-{MNIST}: a novel image dataset for benchmarking machine
  learning algorithms.
\newblock \emph{arXiv preprint arXiv:1708.07747}, 2017.

\bibitem[Yin et~al.(2020)Yin, Farajtabar, Li, Levine, and Mott]{yin2020sola}
Dong Yin, Mehrdad Farajtabar, Ang Li, Nir Levine, and Alex Mott.
\newblock Optimization and generalization of regularization-based continual
  learning: a loss approximation viewpoint.
\newblock \emph{arXiv preprint arXiv:2006.10974}, 2020.

\bibitem[Zenke et~al.(2017)Zenke, Poole, and Ganguli]{zenke2017si}
Friedemann Zenke, Ben Poole, and Surya Ganguli.
\newblock Continual learning through synaptic intelligence.
\newblock In \emph{Proceedings of the 34th International Conference on Machine
  Learning (ICML)}, volume~70 of \emph{PMLR}, pages 3987--3995, 2017.

\end{thebibliography}

\clearpage
\appendix

\section{Numerical checks}\label{app:checks}

Table~\ref{tab:checks} reports, for one pair of tasks at initialisation, the ratio $\|\theta_{AB}-\theta_{BA}\| / (\eta^2\|\br\|)$ for one full step in each order, and the cosine between the measured and predicted gap. Hessian--vector products were checked for symmetry ($u^{\top}Hv$ and $v^{\top}Hu$ agree to within $0.03\%$ on every model).

\begin{table}[h]
\centering\small
\caption{Check of Proposition~\ref{prop:gap}: measured over predicted order gap (cosine in parentheses).}
\label{tab:checks}
\begin{tabular}{@{}lcccc@{}}
\toprule
 & $\eta=0.1$ & $\eta=0.03$ & $\eta=0.01$ & $\eta=0.003$\\
\midrule
Fashion-MNIST (MLP) & 0.878 (0.951) & 0.979 (0.995) & 0.995 (0.999) & 0.999 (1.000)\\
CIFAR-10 features (MLP) & 0.948 (0.996) & 0.984 (1.000) & 0.995 (1.000) & 0.999 (1.000)\\
CIFAR-10 CNN & 1.005 (0.933) & 0.992 (0.994) & 0.996 (0.999) & 0.999 (1.000)\\
\bottomrule
\end{tabular}
\end{table}
\noindent On MNIST the same check gave ratios between $1.001$ and $1.007$ with cosine up to $1.000$.

\section{Learning-rate sweep}\label{app:lr}

\begin{table}[h]
\centering\small
\caption{Split MNIST, three learning rates, 100 runs each. Rank correlation with forgetting. The last column is mean forgetting in accuracy (\%). $^\dagger$Uses old data.}
\label{tab:lr}
\begin{tabular}{@{}llcccccc@{}}
\toprule
 & $\eta$ & Stored bracket & EWC & Taylor & Full bracket$^\dagger$ & Re-eval.$^\dagger$ & Forgetting\\
\midrule
Slow & 0.02 & 0.86 & 0.55 & 0.58 & 0.86 & 0.61 & 1.0\\
Slow & 0.05 & 0.81 & 0.58 & 0.67 & 0.82 & 0.81 & 1.3\\
Slow & 0.1 & 0.71 & 0.59 & 0.79 & 0.73 & 0.84 & 1.2\\
\midrule
Fast & 0.02 & 0.46 & 0.61 & 0.46 & 0.72 & 0.77 & 30.9\\
Fast & 0.05 & 0.40 & 0.59 & 0.41 & 0.69 & 0.81 & 35.0\\
Fast & 0.1 & 0.33 & 0.49 & 0.37 & 0.63 & 0.82 & 37.1\\
\bottomrule
\end{tabular}
\end{table}

On the CIFAR-10 CNN at $\eta=0.02$ the stored bracket improves but stays below EWC: slow $0.22$ vs.\ $0.43$, fast $0.42$ vs.\ $0.54$ (Taylor estimate $0.18$ and $0.80$).

\section{Tiny buffers}\label{app:buffer}

Here the bracket is computed from $n$ stored examples of $A$ instead of the summary, and re-evaluation uses the same $n$ examples.

\begin{table}[h]
\centering\small
\caption{Split MNIST with a buffer of $n$ examples of the old task (100 runs each). ``Br'' is the bracket from the buffer, ``Chk'' is re-evaluation on the buffer.}
\label{tab:buf_mnist}
\begin{tabular}{@{}llcccccccc@{}}
\toprule
 & $\eta$ & EWC & Stored br. & Br 10 & Chk 10 & Br 50 & Chk 50 & Br 200 & Chk 200\\
\midrule
Slow & 0.02 & 0.55 & 0.86 & 0.73 & 0.01 & 0.79 & 0.14 & 0.83 & 0.24\\
Slow & 0.05 & 0.58 & 0.81 & 0.66 & 0.12 & 0.71 & 0.31 & 0.77 & 0.42\\
Slow & 0.1 & 0.59 & 0.71 & 0.57 & 0.11 & 0.65 & 0.32 & 0.69 & 0.59\\
\midrule
Fast & 0.02 & 0.61 & 0.46 & 0.74 & 0.68 & 0.75 & 0.71 & 0.77 & 0.76\\
Fast & 0.05 & 0.59 & 0.40 & 0.72 & 0.71 & 0.72 & 0.79 & 0.72 & 0.81\\
Fast & 0.1 & 0.49 & 0.33 & 0.57 & 0.72 & 0.57 & 0.79 & 0.59 & 0.82\\
\bottomrule
\end{tabular}
\end{table}

\begin{table}[h]
\centering\small
\caption{Buffers on the other setups, $\eta=0.05$ (100 runs each).}
\label{tab:buf_other}
\begin{tabular}{@{}llcccccc@{}}
\toprule
 & & Stored br. & Br 10 & Br 50 & Chk 10 & Chk 50 & EWC\\
\midrule
Fashion-MNIST & slow & 0.70 & 0.41 & 0.52 & 0.29 & 0.29 & 0.33\\
Fashion-MNIST & fast & 0.78 & 0.59 & 0.66 & 0.59 & 0.69 & 0.79\\
CIFAR-10 features & slow & 0.76 & 0.29 & 0.62 & $-$0.35 & $-$0.06 & 0.35\\
CIFAR-10 features & fast & 0.46 & 0.28 & 0.69 & 0.22 & 0.65 & 0.84\\
CIFAR-10 CNN & slow & $-$0.12 & $-$0.12 & $-$0.21 & 0.11 & 0.19 & 0.22\\
CIFAR-10 CNN & fast & 0.30 & 0.12 & 0.28 & 0.48 & 0.64 & 0.47\\
\bottomrule
\end{tabular}
\end{table}

In the slow scenario the data-free stored bracket is better than a bracket computed from 10 or 50 examples on every well-learned setup, and re-evaluation on small buffers is unreliable.

\section{Attempted repairs for fast forgetting}\label{app:fixes}

\begin{table}[h]
\centering\small
\caption{CIFAR-10 features, fast forgetting, $\eta=0.05$. Combinations add the ranks of the signals. Rows with $K=50,100$ use 3 seeds (60 runs); the others use 5 seeds.}
\label{tab:fixes}
\begin{tabular}{@{}lc@{}}
\toprule
Signal (no old data unless stated) & $\rho$ with forgetting\\
\midrule
EWC penalty & 0.84\\
Stored bracket, $K=20$ & 0.46\\
Stored bracket $+$ EWC penalty & 0.71\\
Stored bracket $+$ Taylor estimate & 0.62\\
Stored bracket $+$ EWC penalty $+$ Taylor estimate & 0.75\\
Stored bracket, $K=50$ & 0.49\\
Stored bracket, $K=100$ & 0.49\\
Bracket from 50 examples $+$ EWC penalty & 0.81\\
\bottomrule
\end{tabular}
\end{table}

With $K=50$ and $K=100$ the Taylor estimate rises slightly ($0.76$, $0.77$) and EWC is unchanged ($0.85$ on the same 60 runs).
On Split MNIST, scoring the stored bracket at steps 5, 10, 15 and 20 separately gave $0.45$, $0.38$, $0.36$ and $0.37$, so measuring earlier does not help either.

\section{Diagnostics on the half-learned CNN}\label{app:diag}

Before identifying the residual gradient as the cause, I tested other explanations on the CIFAR-10 CNN ($\eta=0.05$, 3 seeds; slow / fast).
Restricting the bracket to the convolutional layers ($0.00$ / $0.25$) or to the dense layers ($0.05$ / $0.36$) does not help.
Projecting the bracket onto $A$'s gradient gives $0.09$ / $0.41$ with old data and $0.02$ / $0.49$ from the summary, a partial improvement in the fast scenario that remains below EWC ($0.26$ / $0.54$ on the same runs).
Training for 6 epochs instead of 3 raises accuracy only from $0.78$ to $0.79$ and lowers the full bracket's correlation ($-0.09$ / $0.15$).

\section{Curvature at the old task's solution}\label{app:valley}

\begin{table}[h]
\centering\small
\caption{Extreme eigenvalues of $H_A(\thA)$ and alignment of $B$'s gradient with $A$'s curvature (fast scenario, $\eta=0.05$, 2 seeds).}
\label{tab:valley}
\begin{tabular}{@{}lcccc@{}}
\toprule
 & MNIST & Fashion-MNIST & CIFAR-10 features & \cnnhead\\
\midrule
Largest eigenvalue $\lambda_{\max}$ & 5.43 & 3.85 & 2.35 & 16.26\\
Smallest eigenvalue $\lambda_{\min}$ & $-0.30$ & $-0.23$ & $-0.12$ & $-2.21$\\
$|\lambda_{\min}|/\lambda_{\max}$ & 0.065 & 0.072 & 0.051 & 0.150\\
$\cos(g_B,\,H_A g_B)$ & 0.59 & 0.62 & 0.59 & 0.82\\
\bottomrule
\end{tabular}
\end{table}

In the slow scenario the eigenvalue ratios are similar ($0.063$, $0.061$, $0.054$, $0.153$). The cosine is low there on every setup ($0.06$--$0.11$) because most of $g_B$ lies on $B$'s own output weights, which $L_A$ does not depend on.

\section{Five-task sequences: details}\label{app:seq}

Table~\ref{tab:seq_age} gives every signal by age, and Table~\ref{tab:drift} the drift of old tasks.
Measured as totals since the old task ended rather than growth during the new task, the EWC penalty reaches $0.26$, $-0.02$ and $0.22$, and the Taylor estimate $0.35$, $0.03$ and $0.26$, on MNIST, Fashion-MNIST and CIFAR-10 features in the slow scenario, below the growth versions used in the main text in every case.

\begin{figure}[h]
\centering
\includegraphics[width=\linewidth]{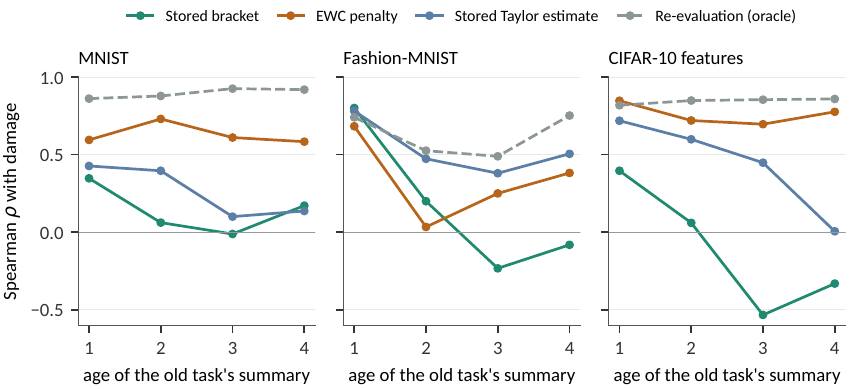}
\caption{Five-task sequences, fast forgetting (Domain-IL), by age of the stored summary.}
\label{fig:seq_fast}
\end{figure}

\begin{table}[h]
\centering\small
\caption{Five-task sequences by age of the old task's summary (number of tasks learned since it was stored). Events per age: 80, 60, 40, 20. $^\dagger$Uses old data.}
\label{tab:seq_age}
\begin{tabular}{@{}llcccccccc@{}}
\toprule
 & & \multicolumn{4}{c}{Slow (Task-IL)} & \multicolumn{4}{c}{Fast (Domain-IL)}\\
\cmidrule(lr){3-6}\cmidrule(l){7-10}
 & Signal & 1 & 2 & 3 & 4 & 1 & 2 & 3 & 4\\
\midrule
MNIST & Stored bracket (proposed) & 0.76 & 0.43 & 0.40 & 0.54 & 0.35 & 0.06 & $-$0.01 & 0.17\\
 & EWC penalty & 0.66 & 0.45 & 0.16 & $-$0.02 & 0.59 & 0.73 & 0.61 & 0.58\\
 & Stored Taylor estimate & 0.79 & 0.70 & 0.63 & 0.70 & 0.43 & 0.40 & 0.10 & 0.14\\
 & Full bracket$^\dagger$ & 0.78 & 0.49 & 0.53 & 0.54 & 0.73 & 0.15 & $-$0.01 & 0.06\\
 & Re-evaluation$^\dagger$ & 0.84 & 0.84 & 0.66 & 0.66 & 0.86 & 0.88 & 0.93 & 0.92\\
\addlinespace
Fashion & Stored bracket (proposed) & 0.66 & 0.35 & 0.15 & $-$0.50 & 0.80 & 0.20 & $-$0.23 & $-$0.08\\
 & EWC penalty & 0.18 & 0.29 & 0.29 & 0.16 & 0.68 & 0.03 & 0.25 & 0.38\\
 & Stored Taylor estimate & 0.40 & 0.68 & 0.46 & 0.55 & 0.78 & 0.47 & 0.38 & 0.51\\
 & Full bracket$^\dagger$ & 0.73 & 0.14 & 0.08 & $-$0.53 & 0.61 & 0.01 & $-$0.10 & $-$0.08\\
 & Re-evaluation$^\dagger$ & 0.45 & 0.69 & 0.50 & 0.67 & 0.74 & 0.53 & 0.49 & 0.75\\
\addlinespace
CIFAR feat. & Stored bracket (proposed) & 0.73 & 0.56 & 0.72 & 0.73 & 0.40 & 0.06 & $-$0.53 & $-$0.33\\
 & EWC penalty & 0.39 & 0.60 & 0.61 & 0.74 & 0.85 & 0.72 & 0.70 & 0.78\\
 & Stored Taylor estimate & 0.51 & 0.58 & 0.46 & 0.47 & 0.72 & 0.60 & 0.45 & 0.01\\
 & Full bracket$^\dagger$ & 0.73 & 0.55 & 0.68 & 0.70 & 0.75 & 0.39 & $-$0.19 & $-$0.03\\
 & Re-evaluation$^\dagger$ & 0.57 & 0.64 & 0.44 & 0.49 & 0.82 & 0.85 & 0.85 & 0.86\\
\bottomrule
\end{tabular}
\end{table}

\begin{table}[h]
\centering\small
\caption{Drift of old tasks in the slow (Task-IL) sequences: mean rise in an old task's test loss since it was learned, measured just before the new task starts, and the mean further rise caused by the new task.}
\label{tab:drift}
\begin{tabular}{@{}lcccccc@{}}
\toprule
 & \multicolumn{3}{c}{Drift before the new task} & \multicolumn{3}{c}{Damage by the new task}\\
\cmidrule(lr){2-4}\cmidrule(l){5-7}
Age & MNIST & Fashion & CIFAR feat. & MNIST & Fashion & CIFAR feat.\\
\midrule
1 & 0.000 & 0.000 & 0.000 & 0.029 & 0.023 & 0.016\\
2 & 0.033 & 0.026 & 0.018 & 0.034 & 0.024 & 0.013\\
3 & 0.060 & 0.047 & 0.030 & 0.041 & 0.017 & 0.020\\
4 & 0.101 & 0.087 & 0.046 & 0.034 & $-$0.004 & 0.016\\
\bottomrule
\end{tabular}
\end{table}

\end{document}